\documentclass[10pt,twocolumn,letterpaper]{article}

\usepackage{cvpr}              
\definecolor{cvprblue}{rgb}{0.21,0.49,0.74}
\usepackage[pagebackref,breaklinks,colorlinks,allcolors=cvprblue]{hyperref}
\usepackage{multirow}
\usepackage{gensymb}
\usepackage{comment}
\usepackage{graphicx}
\usepackage{amsmath, amssymb}

\usepackage{graphicx}  

\usepackage{makecell}

\usepackage{caption}
\usepackage{booktabs}           
\usepackage{float}
\usepackage{amsfonts}
\usepackage{xcolor}
\usepackage{mathtools}
\usepackage[noend]{algpseudocode}

\newcommand{\B}[0]{\ensuremath{\mathcal{B}}}
\newcommand{\A}[0]{\ensuremath{\mathcal{A}}}

\def\paperID{****} 
\def\confName{CVPR}
\def\confYear{2026}

\title{MV-RoMa: From Pairwise Matching into Multi-View Track Reconstruction}

\author{Jongmin Lee$^{*{\dagger}}$\\
KAIST\\
{\tt\small jongmin.cv@gmail.com}
\and
Seungyeop Kang$^{*}$ \\
Seoul National University\\
{\tt\small duq8653@gmail.com}
\and
Sungjoo Yoo\\
Seoul National University\\
{\tt\small sungjoo.yoo@gmail.com}
}

\begin{document}
\maketitle
\renewcommand{\thefootnote}{\fnsymbol{footnote}}
\footnotetext[1]{Equal contribution}
\footnotetext[2]{Work performed at Seoul National University}

\begin{abstract}

Establishing consistent correspondences across images is essential for 3D vision tasks such as structure-from-motion (SfM), yet most existing matchers operate in a pairwise manner, often producing fragmented and geometrically inconsistent tracks when their predictions are chained across views. 
We propose \textbf{MV-RoMa}, a multi-view dense matching model that jointly estimates dense correspondences from a source image to multiple co-visible targets.

Specifically, we design an efficient model architecture which avoids high computational cost of full cross-attention for multi-view feature interaction: (i) multi-view encoder that leverages pair-wise matching results as a geometric prior, and (ii) multi-view matching refiner that refines correspondences using pixel-wise attention.
Additionally, we propose a post-processing strategy that integrates our model's consistent multi-view correspondences as high-quality tracks for SfM.
Across diverse and challenging benchmarks, MV-RoMa produces more reliable correspondences and substantially denser, more accurate 3D reconstructions than existing sparse and dense matching methods. Project page: \url{https://icetea-cv.github.io/mv-roma/}.

\end{abstract}

\section{Introduction}
Feature matching, \ie establishing correspondences between pixels in different images,
is a foundational task for various downstream tasks in computer vision such as 3D reconstruction~\cite{reconstruct_world, build_rome_in_a_day, mast3r-sfm, colmap} and visual localization~\cite{sacreg, visual_loc_2, hloc, onepose}. 
Recently, many 3D vision tasks~\cite{detectorfreesfm, dense-sfm, geoneus, naive_gs} have driven a growing demand for dense matching techniques, as they require dense and reliable correspondences even in textureless or repetitive regions where traditional keypoint-based methods often fail.

However, most existing matching models are fundamentally pairwise (\ie 1-to-1 image matching), which can lead to suboptimal results in multi-view tasks such as SfM~\cite{colmap, build_rome_in_a_day, incrementalsfm, glomap} since chaining pairwise correspondences often yields fragmented and geometrically inconsistent tracks that degrade reconstruction quality.
Existing methods attempt to address this limitation through post-hoc refinement~\cite{detectorfreesfm, dense-sfm}, but such refinement should be performed on each individual track, which is computationally prohibitive for dense matchers and heavily depends on the quality of the initial pairwise matches.

\begin{figure}[t]
    \centering
    \includegraphics[width=0.48\textwidth]{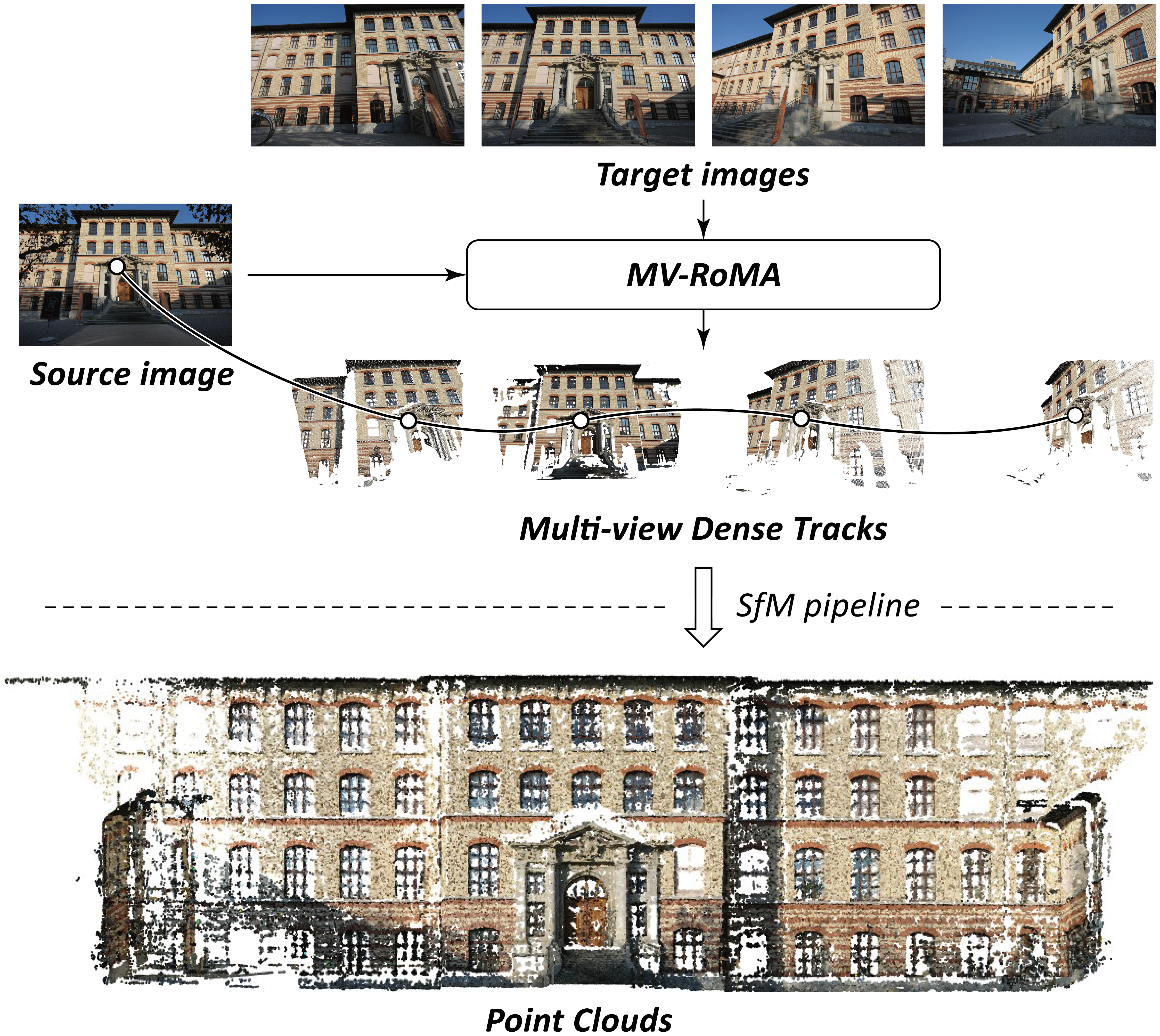}
    \caption{\textbf{Overview of MV-RoMa.} Given a source and multiple co-visible target images, MV-RoMa jointly estimates dense correspondence fields that are geometrically consistent across views (top). These fields are then fed into the SfM pipeline, finally yielding a dense and accurate reconstructed point cloud (bottom).}
    \label{fig:intro}
\vspace{-5mm}
\end{figure}

In this work, we propose 
\textbf{MV-RoMa} (Multi-View Robust dense feature Matching)
that produces consistent tracks by processing multiple images simultaneously, directly addressing these limitations.
To our knowledge, our approach is the first to estimate dense correspondences across multiple views, enabling coherent track reconstruction without the error accumulation inherent in sequential pairwise matching.
The key insight is to embed sparse geometric priors—derived from initial pairwise matches—into the network to guide multi-view feature learning.
Specifically, we perform clustering-based sampling on the prior matches to construct reliable multi-view tracks, which are instantiated as track tokens and integrated into the feature encoder via a track-guided attention mechanism. 
This design enables the encoder to efficiently exchange information along the tracks, producing multi-view-aware representations that aggregate features across views and facilitate accurate correspondence estimation.
We further propose a multi-view matching refiner that estimates fully dense correspondences in a coarse-to-fine manner.
At the final refinement stage, an iterative multi-view attention block warps image features from target views into the source view's coordinate frame and performs pixel-aligned attention, 
thereby avoiding the prohibitive cost of full cross-attention used in prior multi-view models~\cite{mvdust3r, pi3, vggt}. As a result, our model predicts a set of dense correspondences that are jointly estimated across multiple views.

Finally, we introduce a post-processing strategy to integrate the predicted multi-view correspondences into an SfM pipeline~\cite{colmap}.
Since our predictions naturally form 2D tracks across views, we can sample high-quality correspondences that enable dense and accurate 3D reconstruction, as shown in Fig.~\ref{fig:intro}.

The main contributions of this work are summarized as:
\begin{itemize}
    \item A multi-view dense matching framework that directly produces correspondences across multiple images, overcoming the limitations of sequential pairwise matching.
    \item A computationally efficient multi-view feature exchange architecture that consists of the proposed multi-view encoder and matching refiner.
    \item A post-processing strategy that leverages the predicted multi-view-consistent correspondences as reliable tracks within the SfM pipeline.
    \item State-of-the-art results across diverse and challenging benchmarks, consistently outperforming existing matching methods on various downstream tasks.
\end{itemize}

\section{Related Work}

\subsection{Sparse and Dense Pairwise Matching}
Classical image matching pipelines rely on handcrafted keypoints and descriptors~\cite{sift, surf, orb}. To enhance robustness under challenging conditions, early learning-based methods~\cite{superpoint, r2d2, d2net, geodesc, disk, aslfeat, sosnet, s2dnet} follow the \textit{detect--describe--match} paradigm: they first detect keypoints and then extract local descriptors from dense image features. Instead of performing nearest-neighbor matching based solely on descriptor similarity,
recent approaches~\cite{superglue, lightglue, oanet, sgmnet} learn to establish correspondences between keypoints in an end-to-end manner.

Sparse matching methods, however, often struggle in textureless or repetitive regions, where 
obtaining reliable keypoints is challenging and established matches are often ambiguous. 
This limitation has motivated a shift toward (semi-)dense matching approaches~\cite{loftr, efficientloftr, matchformer, aspanformer}, which employ transformer-based architectures with global receptive fields over the entire image to improve correspondence precision and robustness.
To estimate fully dense matches 
for image pairs,
recent works like DKM~\cite{dkm} and RoMa~\cite{roma} 
first predict coarse correspondences and then progressively upsample them to high resolution through iterative refinement. 
More recently, UFM~\cite{ufm} unified wide-baseline matching and optical flow within a single model, demonstrating strong generalization across diverse correspondence tasks.

\subsection{Multi-View Consistency}

The prevailing strategy for multi-view matching is to decompose the problem into pairwise subproblems, merge the pairwise results, and then check multi-view projection consistency~\cite{build_rome_in_a_day, hloc, colmap_localization, colmap}.
However, recent studies~\cite{pixsfm, detectorfreesfm, dense-sfm} have shown that naively aggregating pairwise matches often yields fragmented and inconsistent tracks—particularly when dense matchers are employed.
A common approach is to apply post-processing~\cite{practical_mv_matching, unified_cycle_consistency}, 
which filters and refines tracks by enforcing consistency constraints on the initial correspondences.
Another line of works~\cite{pixsfm, detectorfreesfm, dense-sfm} instead optimizes keypoint locations within tracks to obtain geometrically consistent, multi-view–aligned correspondences.
Although these approaches improve performance, they still fundamentally depend on the initial pairwise matches, which can limit the reliability of the resulting multi-view tracks and introduce additional computational overhead, as each track should be refined individually.

A few recent works have explored joint reasoning across multi-view images. End-to-end multi-view matchers~\cite{comatcher, e2e_mvmatch} introduced an attention-based framework for collaborative sparse matching across co-visible image sets. However, these methods remain constrained to pre-detected keypoints, which limits matching precision due to their dependence on keypoint detectors. 
In contrast, our method, MV-RoMa, jointly reasons over multiple views to predict fully dense correspondences from a source image to multiple target images. As a result, our method produces more geometrically consistent tracks than keypoint-based methods.

\section{Method}
\label{sec:method}

\begin{figure*}[ht!]
    \centering
    \includegraphics[width=\textwidth]{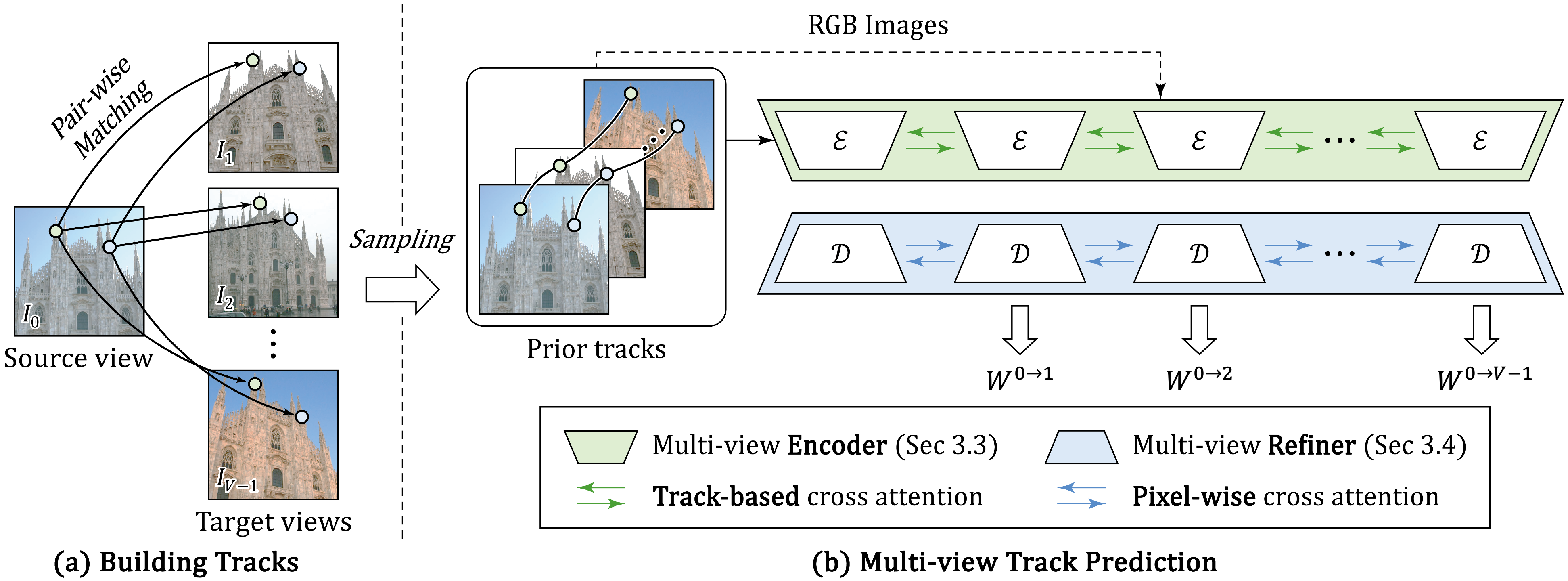}
    \caption{\textbf{Pipeline of MV-RoMa.}
    \textbf{(a) Building Tracks.} Given a source view $I_0$ and a set of target views $\{I_v\}$, we first obtain initial pairwise matches from an off-the-shelf matcher and apply a sampling procedure to construct a sparse set of multi-view \textit{prior tracks}.
    \textbf{(b) Multi-view Track Prediction.} The RGB images and prior tracks are fed into our \textbf{Multi-view Encoder} (Sec.~\ref{sec:method-encoder}), which uses track-based cross-attention to produce geometrically consistent dense features. The \textbf{Multi-view Refiner} (Sec.~\ref{sec:method_decoder}) then applies pixel-wise cross-attention to predict the final dense correspondences $W^{0\rightarrow v}$ for all target views.}
    \label{fig:pipeline}
\vspace{-4mm}
\end{figure*}

\subsection{Overview}
\label{sec:method-overview}
An overview of the pipeline is illustrated in Fig.~\ref{fig:pipeline}. 
Given a source image and a set of co-visible target images, we first perform pairwise matching between the source and each target to obtain initial correspondences, which we aggregate into a sparse geometric prior in the form of multi-view tracks.
Then, this prior is used to guide our track-guided multi-view encoder, producing multi-view-aligned dense feature maps. 
These features are then processed by subsequent modules, including a multi-view matching refiner, to predict dense and geometrically consistent correspondences across all target views.

We first review the coarse-to-fine matching framework on which our method is built (Sec.~\ref{sec:method-prelim-roma}). 
We then describe our core contributions: the multi-view encoder (Sec.~\ref{sec:method-encoder}), which constructs and utilizes the track tokens, and the multi-view refiner (Sec.~\ref{sec:method_decoder}).
Finally, we present a post-processing strategy for integrating the estimated multi-view dense correspondences into an SfM pipeline (Sec.~\ref{sec:method-postprocess}).

\subsection{Preliminary: RoMa}
\label{sec:method-prelim-roma}

\paragraph{Problem setup.}
Given a source image \(I^\A\) and a target image \(I^\B\), dense matching seeks a correspondence for any coordinate \(x^\A\) in \(I^\A\) to a coordinate \(x^\B\) in \(I^\B\).
Following a coarse-to-fine paradigm, RoMa first estimates the coarse correspondences using semantically rich but spatially coarse feature \(\varphi_{\text{coarse}}\) from DINOv2~\cite{dinov2},
and then refines the correspondences using high-resolution feature \(\varphi_{\text{fine}}\) from CNN encoder. Formally,
\begin{equation}
\left\{
\begin{aligned}
    &\big(W^{\A\to\B}_{\text{coarse}},\, p^{\A}_{\text{coarse}}\big)
        = G_{\theta}\big(\varphi^{\A}_{\text{coarse}},\,\varphi^{\B}_{\text{coarse}}\big),\\
    &\big(W^{\A\to\B},\, p^{\A}\big)
        = R_{\theta}\!\Big(W^{\A\to\B}_{\text{coarse}},\, p^{\A}_{\text{coarse}},\,
        \varphi_{\text{fine}}^{\A},\,\varphi_{\text{fine}}^{\B}\Big),
\end{aligned}
\right.
\end{equation}
where \(W^{\A\to\B}_{\text{coarse}}\) and \(p^{\A}_{\text{coarse}}\) represent the coarse warp and confidence from the global matcher \(G_\theta\) respectively, and \(W^{\A\to\B}\), \(p^{\A}\) are the final predictions from the refiner \(R_\theta\).
We adopt the convention that a warp with superscript \(\A\!\to\!\B\) maps coordinates in \(I^\A\) to \(I^\B\).

\paragraph{Global matching.}
The global matcher \(G_\theta\) employs a transformer decoder to predict anchor probabilities over the image grid and estimates
coarse correspondence \(W^{\A\to\B}_{\text{coarse}}\) via a regression-by-classification strategy.

\paragraph{Refinement.}
The refiner \(R_\theta\) consists of a multi-scale stack of ConvNets that progressively refine the correspondence.
Let levels be indexed by \(i \in \{1,\dots,L\}\), where larger \(i\) denotes coarser resolution (stride \(2^{i-1}\)).
At level \(i\), given the previous estimate \((W^{\A\to\B}_{i+1},\, p^{\A}_{i+1})\), 
a local correlation volume is first constructed around the current target location (optional per level), and image feature from \(I^\B\) is warped toward \(I^\A\) following:
\begin{equation}
\left\{
\begin{aligned}
&\mathrm{corr}_{i}
  = \operatorname{corr}\!\big(
       \varphi^{\mathcal{A}}_{fine_{i}},\,
       \varphi^{\mathcal{B}}_{fine_{i}},\,
       W^{\mathcal{A}\to\mathcal{B}}_{i+1}
     \big) \\
&\tilde{\varphi}^{\mathcal{B}}_{fine_{i}}
  = \mathcal{W}\!\big(\varphi^{\mathcal{B}}_{fine_{i}},\, W^{\mathcal{A}\to\mathcal{B}}_{i+1}\big).
\end{aligned}
\right.
\end{equation}
The constructed inputs are aggregated and then processed through ConvNet blocks \(f_i\) to estimate residual updates via output heads \(g_i\) as:
\begin{equation}
\big(\Delta W^{\A\to\B}_{i},\, \Delta p^{\A}_{i}\big) = 
 g_i\big(f_i\!\big(\varphi^{\A}_{i},\, \tilde{\varphi}^{\B}_{i},\, \mathrm{corr}_{i}\big)\big)
\end{equation}
Finally, the correspondence and confidence are updated as:
\begin{equation} 
\left\{
\begin{aligned}
W^{\A\to\B}_{i} &= \operatorname{Up}\!\big(W^{\A\to\B}_{i+1}\big) + \Delta W^{\A\to\B}_{i},\\
p^{\A}_{i} &= \operatorname{Up}\!\big(p^{\A}_{i+1}\big) + \Delta p^{\A}_{i}.
\end{aligned}
\right.
\end{equation}

where \(\operatorname{Up}(\cdot)\) upsamples the previous estimation to level \(i\)'s resolution. Iterative refinement to the finest level \(i{=}1\) yields the final warp \(W^{\A\to\B}\) and confidence \(p^{\A}\).

\subsection{Track-Guided Multi-View Encoder}
\label{sec:method-encoder}


Given a source image $I_0$ and a set of target images $\{I_v\}_{v=1}^{V-1}$, our goal is to extract dense feature maps that are geometrically consistent across all views. 
To enable information exchange across views during feature extraction, we leverage 2D point tracks—which we term \textit{track tokens}—as a sparse geometric prior. 
We construct these track tokens via clustering-based sampling from off-the-shelf pairwise matches. 
We then integrate the track tokens into the encoder through a process inspired by Tracktention~\cite{tracktention}: 
we first \textit{sample} features from the image grids onto the tracks, \textit{propagate} the features along each track across views, and finally \textit{splat} the updated, multi-view-aware features back onto the image grids. This process produces the multi-view aligned feature maps.

\subsubsection{Building Tracks with an Image Matcher}
\label{sec:method-track-construction}

\paragraph{Goal and Motivation.}
Given pairwise correspondences from an off-the-shelf matcher (either sparse, semi-dense, or dense), our goal is to summarize them into a set of~$T$ 
multi-view tracks.
A naive approach, such as random sampling, can result in spatially redundant samples and is sensitive to noisy matches. To address this, we propose a \textit{clustering-based sampling} 
scheme that produces a compact geometric prior for subsequent stages.

\paragraph{Track Token Representation.}
We denote the set of track tokens by $\mathcal{T} = \{t_i\}_{i=1}^{T}$, where each token $t_i$ encapsulates a multi-view correspondence.
It contains a coordinate vector $\mathbf{u}_i \in \mathbb{R}^{2V}$ that stacks the 2D positions across all $V$ views
\[
\mathbf{u}_i = [x_0, y_0, x_1, y_1, \dots, x_{V-1}, y_{V-1}]^\top,
\]
where coordinates for missing views are padded with a sentinel value (\eg with -1). To handle partial tracks that are visible only in a subset of views, each token also includes a binary visibility mask $\mathbf{m}_i \in \{0,1\}^{V}$. This mask indicates whether the track is observable in each view, with a value of 1 for an observed view and 0 for a missing one; the entry for the source view is always 1 by definition. 
During multi-view aggregation, we use $\mathbf{m}_i$ to ensure that only features from valid, visible views contribute to the updates.
A complete track token is thus defined as $t_i = (\mathbf{u}_i, \mathbf{m}_i)$.

\paragraph{Track Sampling.}
Our track sampling procedure consists of three steps. 
First, we run a pairwise matcher between the source image and each target image to obtain 
an initial set of multi-view tracks.
Second, we group these tracks into partitions based on their visibility masks, so that all tracks within a group are visible in the same subset of images. 
Finally, within each partition, we construct track tokens by using $k$-means clustering~\cite{kmeans_clustering} to select representative matches.
Specifically, the number of clusters assigned to each partition is proportional to its size, and the total number of clusters across all partitions equals the desired number of tokens $T$.
The representative track token for each cluster is then defined by selecting the raw track 
whose coordinate vector is closest to the cluster centroid.
This process yields the final set of track tokens $\mathcal{T}=\{t_i\}_{i=1}^{T}$.

\subsubsection{Track-Guided Feature Extractor}
\label{sec:method-track-integration}
We adopt DINOv2~\cite{dinov2} as our backbone encoder, which is designed to process single images independently without view-consistency.
To inject multi-view context into this encoder, we insert a track-guided module into each transformer block in the second half of DINOv2.
Inspired by the Tracktention architecture~\cite{tracktention}, this module implements a three-stage procedure to exchange information across views:
(i) attentional sampling from the image grids onto the track tokens, 
(ii) multi-view information propagation along each track, and (iii) attentional splatting of the updated track features back onto the image grids.

\paragraph{Attentional Sampling.}
The first stage gathers information from each view's feature map $F_v \in \mathbb{R}^{HW \times D_f}$ onto the sparse track tokens. 
For each view~$v$, we perform cross-attention where the query features are derived from the 2D track coordinates $P_v \in \mathbb{R}^{T \times 2}$ via a small MLP, and the keys and values are given by the image-grid features $F_v$:

\begin{equation}
    Q_v = \text{MLP}(P_v), \quad K_v = V_v = F_v.
\end{equation}
The resulting track features $Z_v$ are then obtained via attention, where we incorporate a spatial bias $B_v$ to favor nearby grid tokens:
\begin{equation}
\label{eq:attentional_sampling}
    Z_v = \mathrm{softmax}\left( \frac{Q_v K_v^\top}{\sqrt{D}} + B_v\right) V_v.
\end{equation}
Here, $B_v \in \mathbb{R}^{T \times HW}$ is defined as a negative squared Euclidean distance,
\[
[B_v]_{ij} = -\frac{\|\mathbf{p}_{vi} - \mathrm{pos}(F_{vj})\|^2}{2\sigma^2},
\]
where $\mathbf{p}_{vi}$ is the 2D coordinate of track $i$ in view $v$, and $\mathrm{pos}(F_{vj})$ denotes the 2D center of grid token $j$. The hyperparameter $\sigma$ is a fixed spatial scale that controls the strength of the locality bias.

\paragraph{Track Transformer.}
Next, we aggregate information across views for each track. For each track $i$, a lightweight transformer 
performs self-attention on the features $\{Z_{v,i}\}_{v=0}^{V-1}$ along the view axis,
yielding updated track features $Z'_{v,i}$. In contrast to Tracktention~\cite{tracktention}, this self-attention mechanism is view-order invariant, as we do not introduce any view-index embeddings. We further apply the visibility mask $\mathbf{m}_i$ from the track token to the attention weights, ensuring that only observed views contribute to the feature aggregation for each track.

\paragraph{Attentional Splatting.}
In the final stage, we splat the multi-view-aware track features $Z'_v$ back onto each view's image-grid features. 
This is achieved by performing the operation in the opposite direction to the sampling stage:
the image grid coordinates now serve as queries, while the updated track features $Z'_{v,i}$ serve as keys and values. As in the track transformer, we apply the visibility masks (denoted as $M'_v$ in the attention score) to ensure that only observable tracks contribute to the image features. This yields the multi-view-consistent feature map $\widetilde{F}_v$:
\begin{equation}
\label{eq:attentional_splatting}
    \widetilde{F}_v = F_v + \mathrm{softmax}\left(\frac{Q'_v {K'_v}^{\!\top}}{\sqrt{D}} + B_v + {M'_v}\right) V'_v W_{\text{out}},
\end{equation}
where $Q'_v$ is derived from the image-grid coordinates, $K'_v$ and $V'_v$ are projected from the updated track features $Z'_v$, and $B_v$ is the same spatial bias as in Eq.~\eqref{eq:attentional_sampling}.

\subsection{Multi-view Matching Refiner}
\label{sec:method_decoder}

We follow the coarse-to-fine pipeline of RoMa~\cite{roma} to estimate fully dense correspondences.
First, we leverage the rich geometric context from our multi-view encoder to estimate coarse correspondences at a low resolution.
Then, we employ a refinement network that takes fine-scale features from a VGG19 backbone~\cite{vgg} and progressively upsamples the coarse matches to full resolution.

\paragraph{Coarse Global Matching.}
To compute initial coarse correspondences for each source-target pair $(I_0, I_v)$, we feed the multi-view-aligned features $\widetilde{F}(\cdot)$ from our encoder into the global matcher~$G$ of RoMa:
\begin{equation}
    W^{0\rightarrow v}_{\text{coarse}} \leftarrow G\big(\widetilde{F}(I_0),\, \widetilde{F}(I_v)\big).
\end{equation}

\begin{figure}[t]
    \centering
    \includegraphics[width=0.48\textwidth]{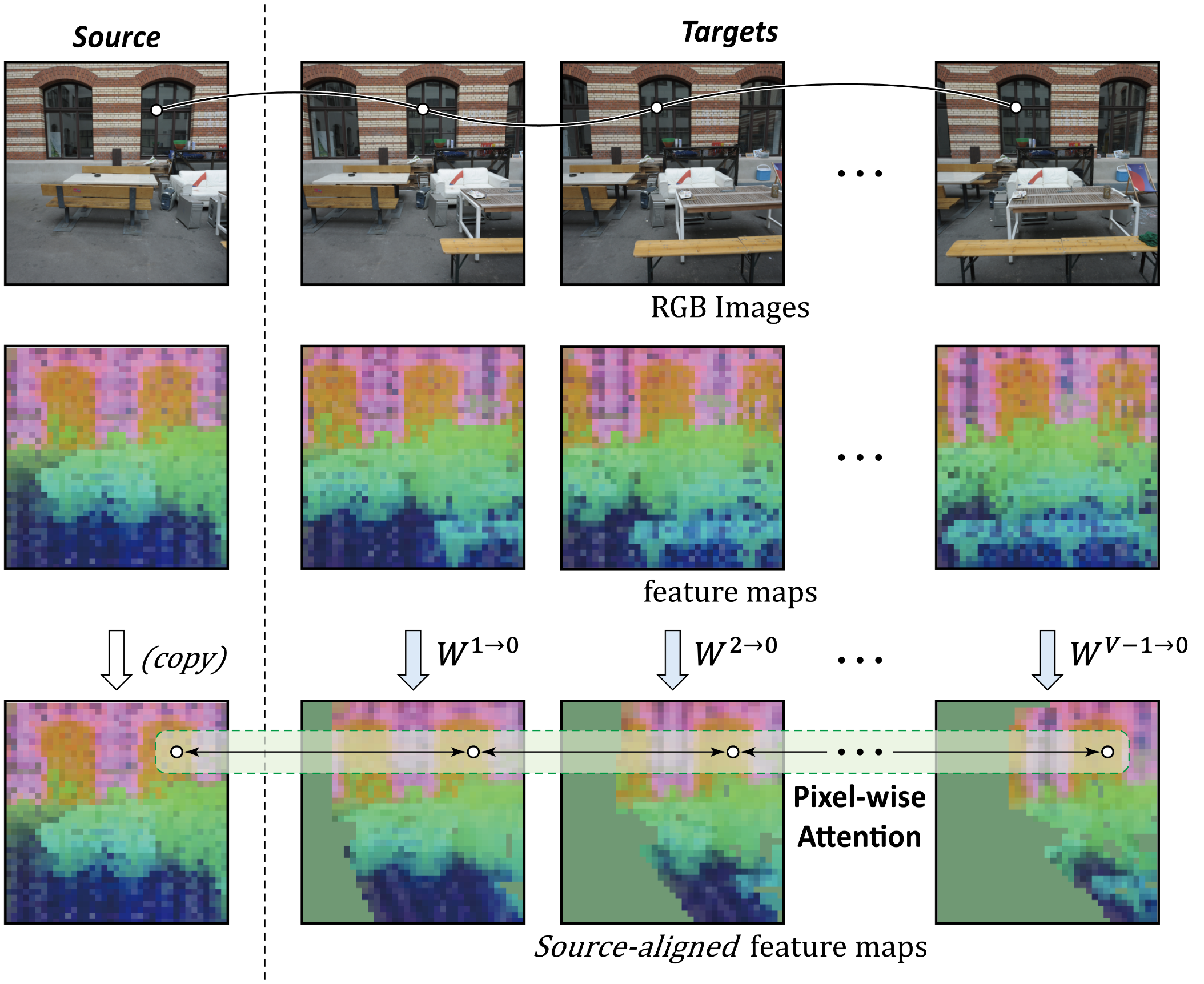}
    \caption{\textbf{Pixel-aligned multi-view attention.} Target features are warped to the source grid using $W^{v\rightarrow 0}$, and per-pixel attention is performed across the aligned views. This avoids the quadratic cost of global cross-attention while refining fine-grained correspondence estimates (Sec.~\ref{sec:method_decoder}).}
    \label{fig:refiner}
\vspace{-5mm}
\end{figure}

\paragraph{Multi-view Matching Refinement.}
We then refine the coarse warps $ W^{0\rightarrow v}_{\text{coarse}}$ using a feature pyramid with strides $s \in \{8, 4, 2, 1\}$ from the VGG19 backbone, following the iterative refinement process described in Sec.~\ref{sec:method-prelim-roma}. While RoMa's refinement relies solely on single-view fine features, our main contribution is to inject multi-view context at the refinement stage.
To achieve this efficiently, as illustrated in Fig.~\ref{fig:refiner}, we spatially align each target view's feature map to the source view using the previous stride's warp estimate.
This alignment enables an efficient \textit{pixel-aligned attention} mechanism, in which each pixel's feature attends to its geometrically corresponding pixels across views. 
In this way, we avoid computationally expensive global attention while still enabling multi-view feature fusion.

Specifically, at level $i$ $\in \{4,1\}$ (\ie at stride \(2^{i-1}\)), 
we apply a stack of $N$ multi-view attention blocks. 
This block alternates between (i) per-pixel feature aggregation across all aligned views via attention and (ii) lightweight spatial propagation within each view using a ConvNeXt block~\cite{convnext}.
The resulting multi-view feature is added as a residual to the refinement hidden state $h_i$ at each iteration $n$:

\begin{equation}
\left\{
    \begin{aligned}
        h^{(0)}_{i} &= f_{i}\big(\varphi^{0}_{i}, \{\tilde{\varphi}^{v}_{i}\}_{v=1}^{V-1}\big), \\
        h^{(n)}_{i} &= \mathrm{MVFuse}^{(n)}\big(h^{(n-1)}_{i}\big).
    \end{aligned}
\right.
\end{equation}

After $N$ iterations, the final warp residual is computed from the updated hidden state:
\begin{equation}
    \big(\Delta W^{0\rightarrow v}_{i}, \Delta p^{0\rightarrow v}_{i}\big) = g_{i}\big(h^{(N)}_{i}\big).
\end{equation}
Here, $\mathrm{MVFuse}^{(n)}(\cdot)$ denotes the $n$-th multi-view fusion step (pixel-aligned attention followed by a ConvNeXt block) operating on the source features $\varphi^{0}_{i}$ and the source-aligned target features $\{\tilde{\varphi}^{v}_{i}\}_{v=1}^{V-1}$.

\subsection{Post-Processing for Structure-from-Motion}
\label{sec:method-postprocess}
In this section, we describe how we integrate our multi-view dense correspondences into an SfM pipeline. In typical SfM pipelines, multi-view tracks are obtained by chaining the outputs of pairwise matchers into longer tracks, whereas our approach directly predicts multi-view correspondences, eliminating the need for explicit pairwise merging.

To integrate these results into an SfM pipeline, we first divide the full image set into groups, each consisting of one source image and $N$ target images, and run our matcher independently on each group. 
The resulting dense multi-view correspondences are then processed via a two-stage procedure consisting of (i) match selection and (ii) track sampling, producing robust tracks suitable for SfM reconstruction. Details of the image grouping strategy are provided in the supplementary material.

\begin{figure}[t!]
    \centering
    \includegraphics[width=0.48\textwidth]{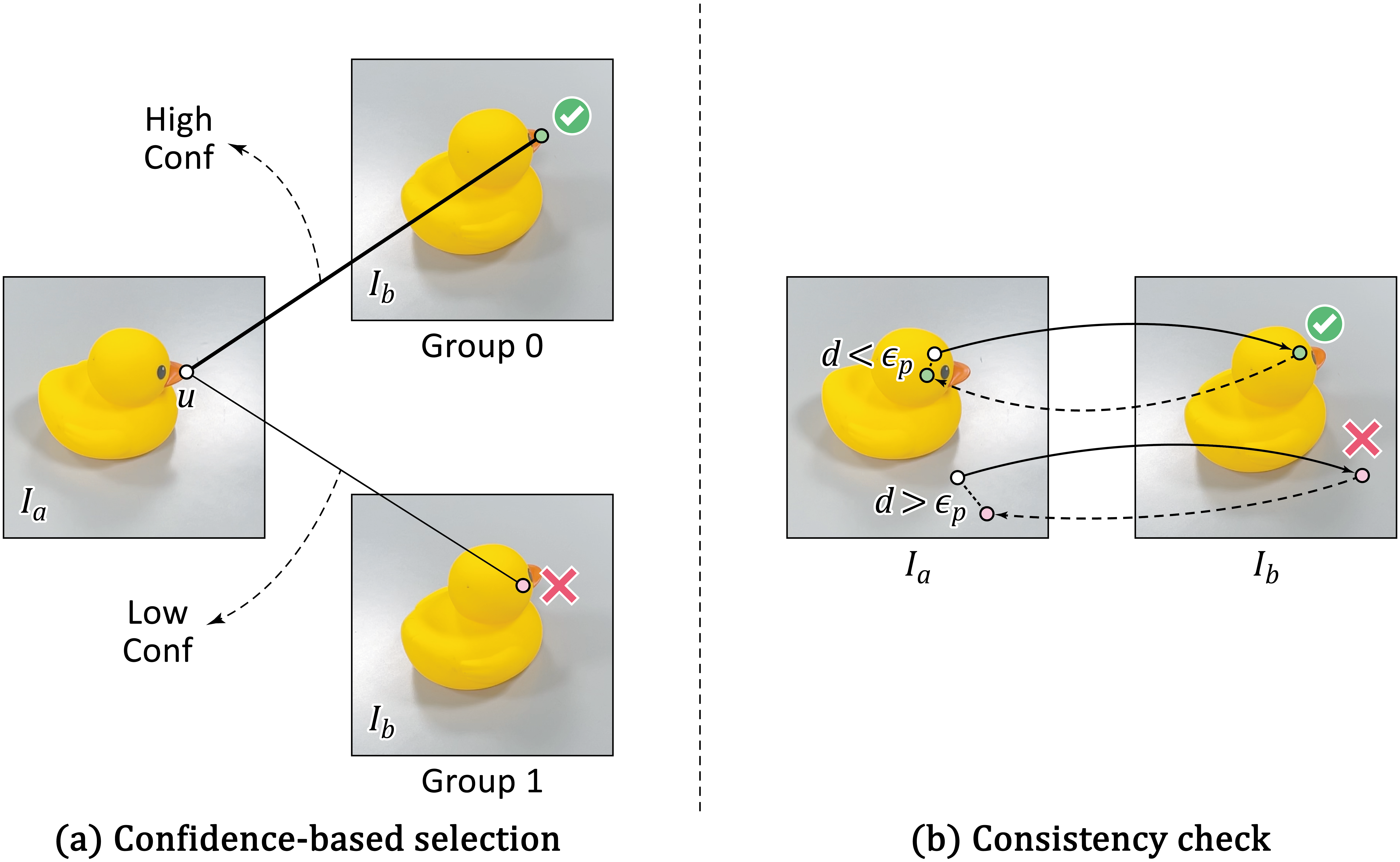}
    \caption{\textbf{Confidence selection and reciprocity filtering.} 
    \textbf{(a)} For each pixel $u$, we select the correspondence with the highest confidence $p_{*}^{a\rightarrow b}(u)$ from multiple predictions across groups.
    \textbf{(b)} We then apply bidirectional (forward–backward) consistency filtering, retaining matches whose cycle error is below the threshold $\epsilon_p$ (Sec.~\ref{sec:method-postprocess}).}
    \label{fig:reprocity}
\vspace{-5mm}
\end{figure}

\subsubsection{Match Selection and Filtering}
\label{pooling_mutual}

This stage processes matches for each ordered image pair $(I_a, I_b)$, treating $I_a$ as the source and $I_b$ as the target. 
Let $\mathcal{G}$ denote the set of all image groups, and let $\mathcal{G}_{a,b} \subseteq \mathcal{G}$ be the subset of groups in which correspondences $W^{a \rightarrow b}(u)$ is available for the pair $(I_a, I_b)$. 
Since the same pair can appear in multiple groups,
a pixel $u$ in $I_a$ may have multiple candidate correspondences in $I_b$.
To handle multiple correspondences from the same source pixel, we select the one with the highest confidence (Fig.~\ref{fig:reprocity}.a).
Specifically, for each pixel $u$ we find the group $g_*$ 
that maximizes the confidence $p^{a \rightarrow b}_{(g)}(u)$:
\begin{equation}
    g_*(u) = \underset{g \in \mathcal{G}_{a,b}}{\arg\max} \, p^{a \rightarrow b}_{(g)}(u).
\end{equation}
The final correspondence $W_{*}^{a \rightarrow b}(u)$ and its confidence $p_{*}^{a \rightarrow b}(u)$ are then taken from this optimal group $g_*(u)$.

After this step, we apply a bidirectional consistency check to further ensure the reliability of the correspondences~\cite{dense-sfm} (Fig.~\ref{fig:reprocity}.b). We retain a correspondence only if its forward-backward cycle error is below a threshold $\epsilon_p$:
\begin{equation}
    \big\| u - W_{*}^{b \rightarrow a}\big(W_{*}^{a \rightarrow b}(u)\big) \big\|_2 \le \epsilon_p.
\label{eq:mutual}
\end{equation}

\subsubsection{Track Sampling and Construction}
\label{nms}
To ensure computational efficiency in the SfM pipeline, we sample a subset of the filtered correspondences to construct the final tracks.
Within each image group, we aggregate the matching confidence from all source-to-target pairs to compute a score map $S \in \mathbb{R}^{H\times W}$.
The score $S(u)$ for each pixel $u$ is designed to reward both the length and the quality of the potential track originating from that pixel. We define two metrics: the \textit{track length} $L(u)$, given by the number of target views with a valid correspondence (confidence $> \tau$), and the \textit{average match confidence} $C(u)$ over those valid views. The final score combines these two metrics as:
\begin{equation}
    S(u) = L(u) + C(u).
\end{equation}
Finally, we apply non-maximum suppression (NMS)~\cite{superpoint, superglue} to the score map~$S$ to select the final keypoints on the source image. This sampling strategy improves
both the efficiency and the stability of the SfM pipeline (\eg bundle adjustment~\cite{bundle_adjustment}).
Each selected keypoint on the source image, 
together with its valid correspondences in the target views, then forms a final multi-view track.

\section{Experiments}

\subsection{Implementation Details.}
\paragraph{Training setup.}
We train our model for 200K steps with a batch size of 4, using the robust loss function from RoMa~\cite{roma}. The learning rate is initialized to $3 \times 10^{-5}$ and decayed by a factor of 10 after 20K steps. Our training data combines sampled pairs from MegaDepth~\cite{megadepth} and ScanNet~\cite{scannet} datasets, excluding scenes reserved for testing (\ie the IMC 2021 PhotoTourism dataset~\cite{imc_dataset}). 
Supervision signals (warps and co-visible areas) are derived from dense depth maps: for MegaDepth, depth is obtained via multi-view stereo (MVS), and for ScanNet, we use the depth maps provided by the RGB-D sensor. 
Details on how we construct multi-view training groups for each dataset are provided in the supplementary material.

\paragraph{Model and Pipeline Configuration.}

By default, we set the number of track tokens to $T=512$. For all ablation studies and evaluations, track tokens are constructed using the UFM~\cite{ufm} dense matcher. Unless otherwise specified, our matcher operates on image groups consisting of one source and four target images. 
During inference, the base resolution is set to $672\times672$, and predictions are upsampled via the refiner following prior work~\cite{roma}. Specifically, we upsample to $1344\times1344$ for HPatches and IMC dataset evaluations, to $1200\times800$ for ETH3D dataset, and to $840 \times 840$ for the Texture-Poor SfM dataset. Ablation studies are conducted without upsampling.
In the SfM post-processing stage, we generate multi-view tracks from the matcher outputs via a three-stage procedure: (i) bidirectional consistency check with a threshold of $\epsilon_p=3$ (Eq.~\ref{eq:mutual}), (ii) confidence-based filtering to remove invalid matches with a threshold of $\tau=0.3$, and (iii) non-maximum suppression with a 2-pixel radius.

\subsection{Homography Estimation}
\paragraph{Evaluation Protocol.}

We evaluate our matcher on the HPatches dataset~\cite{hpatches}, which contains image sequences with changes in illumination and viewpoint. Each scene consists of one source image and five target images, forming five source–target pairs per scene. For each pair, we estimate a homography from the predicted correspondences using either a direct linear transform (DLT) solver~\cite{multi_view_geometry} or a robust estimator (RANSAC)~\cite{ransac}. 
For each image pair, we compute the mean reprojection error at the four image corners and evaluate performance using the area under the cumulative error curve (AUC).
We follow the evaluation protocol proposed by LoFTR~\cite{loftr} and RoMa~\cite{roma}.
Dense matching methods use balanced sampling~\cite{dkm} to select a comparable number of correspondences. 
Our model, MV\mbox{-}RoMa, produces matches to all five target images in a single forward pass, whereas baselines process each pair independently.

 \begin{table}[t!]
    \centering
    \caption{\textbf{Homography estimation on HPatches.} We report the area under the cumulative error curve (AUC) up to values of 1px, 3px and 5px, using DLT and RANSAC~\cite{ransac} as a homography solver.}
    \label{tab:tab_homography}
    \renewcommand{\arraystretch}{1.15}
    \setlength{\tabcolsep}{3pt}
    \small

\resizebox{0.95\columnwidth}{!}{
  \begin{tabular}{ccc} 
    \toprule
    \multirow{2}{*}{Method} & AUC - DLT & AUC - RANSAC \\ 
    \cmidrule(lr){2-2} \cmidrule(lr){3-3}
    & \multicolumn{2}{c}{@1px / @3px / @5px} \\ 
    \midrule
    ALIKED~\cite{aliked}+LG~\cite{lightglue} & 32.6 / 64.6 / 75.3 & 28.4 / 61.5 / 73.8 \\
    LoFTR~\cite{loftr} & 30.2 / 53.3 / 61.6 & 38.0 / 66.9 / 76.2 \\
    Mast3r~\cite{mast3r} & 25.7 / 52.9 / 64.3 & 37.1 / 63.4 / 73.7 \\
    UFM~\cite{ufm} & 27.4 / 58.8 / 70.8 & 29.2 / 61.3 / 73.2 \\
    RoMa~\cite{roma} & 41.0 / 67.9 / 76.9 &  44.7 / 72.6 / 81.4 \\
    MV-RoMa & \textbf{46.1 / 71.9 / 80.1} &  \textbf{47.2 / 73.2 / 81.8} \\
    \bottomrule
    \end{tabular}
    }

\vspace{-3mm}

\end{table}

\paragraph{Results.}
Tab.~\ref{tab:tab_homography} reports AUC@1/3/5\,px for both DLT and RANSAC. MV\mbox{-}RoMa achieves the best performance across all thresholds and both estimators, with the largest gap at the strict 1\,px setting. The small gap between DLT and RANSAC results suggests high inlier matching quality.

\subsection{3D Triangulation}
\paragraph{Evaluation Protocol.}
We evaluate multi-view correspondence quality on the training split of the ETH3D dataset~\cite{eth3d_dataset} by triangulating 3D points from the predicted tracks using a COLMAP pipeline~\cite{colmap} with fixed ground-truth intrinsics and extrinsics. The dataset contains 13 indoor and outdoor scenes with sparsely captured high-resolution images and ground-truth camera poses calibrated using LiDAR. 
Following prior work~\cite{detectorfreesfm,dense-sfm}, we report \textit{Accuracy} and \textit{Completeness}, defined as the proportion of triangulated points that fall within the specified distance thresholds from the ground-truth points.
For baseline pairwise matchers, we decompose each multi-view image group into individual source-target pairs to construct a matching pair dataset, and run the matchers on each pair independently.

 \begin{table}[t!]
    \centering
    \centering

    \caption{\textbf{Results of 3D Triangulation.} Our method is compared with the baselines on the ETH3D~\cite{eth3d_dataset} dataset using accuracy and completeness metrics with different thresholds.}
    \label{tab:exptriangulation}

    \resizebox{\columnwidth}{!}{
      \setlength\tabcolsep{6pt}
      \begin{tabular}{ccccccc} 
        \toprule
        \multirow{2}{*}{Method}         & \multicolumn{3}{c}{Accuracy~($\%$)}          & \multicolumn{3}{c}{Completeness~($\%$)} \\ 
        \cmidrule(lr){2-4}
        \cmidrule(lr){5-7}
                              & 1cm     & 2cm     & 5cm     & 1cm     & 2cm     &  5cm \\ 
        \midrule
        \multicolumn{7}{l}{\textit{Type A. Including Two-view tracks}} \\ 
        \midrule
        
        ALIKED + LG & 55.11 & 69.63 & 84.94 & 0.32 & 1.41 & 6.56\\
        LoFTR & 35.90 & 53.50 & 75.00 & 1.48 & 6.03 & 21.63\\
        Mast3R & 38.99 & 57.2 & 79.16 & 1.44 & 6.83 & 29.83\\
        UFM & 41.55 & 60.07 & 80.44 & 1.12 & 5.19 & 21.34 \\     
        RoMa & \underline{75.58} & \underline{86.25} & \underline{94.95} & \underline{5.64} & \underline{15.73} & \underline{38.60} \\
        \textbf{MV-RoMa} & \textbf{81.91} & \textbf{90.45} & \textbf{96.93} & \textbf{7.52} & \textbf{18.50} & \textbf{41.22} \\
        \midrule
        \multicolumn{7}{l}{\textit{Type B. Excluding Two-view tracks}} \\
        \midrule
        RoMa + DF-SfM & 79.32 & 88.42 & 95.82 & 3.13 & 9.79 & \underline{29.10} \\
        RoMa + Dense-SfM &  \underline{84.79} & \underline{92.62} & \underline{97.77} & \textbf{7.38} & \textbf{17.06} & \textbf{36.35} \\ 
        \textbf{MV-RoMa} &  \textbf{85.88}  & \textbf{92.99} & \textbf{98.05} & \underline{3.95} & \underline{9.94} & 23.81 \\ 
        \bottomrule
      \end{tabular}
    }
\vspace{-3mm}    
\end{table}

\paragraph{Results.}
As shown in Tab.~\ref{tab:exptriangulation}, MV\mbox{-}RoMa achieves the best accuracy and completeness among all pairwise matchers (\textit{Type A}). When compared against full SfM pipelines (\textit{Type B}), MV\mbox{-}RoMa achieves the highest accuracy while remaining competitive in completeness.

\subsection{Multi-View Camera Pose Estimation}
\paragraph{Evaluation Protocol.}

We evaluate multi-view camera pose estimation on two benchmarks: the Texture-Poor SfM dataset~\cite{detectorfreesfm} and the IMC 2021 PhotoTourism dataset~\cite{imc_dataset}. 
The Texture-Poor SfM dataset focuses on object-centric scenes with low texture and sparse viewpoints.
For IMC PhotoTourism, we use all test scenes with subsampled image sets. In both benchmarks, the main challenge lies in significant viewpoint and illumination changes across sparse views.
Following prior work~\cite{detectorfreesfm,pixsfm}, we convert the recovered multi-view camera set into pairwise relative poses and compute the \emph{pose error} for each pair as the maximum of the angular rotation and translation errors. We then report the area under the curve (AUC) of the cumulative pose-error distribution at varying thresholds, averaged over test scenes. 

\paragraph{Results.}

As shown in Tab.~\ref{tab:exp_camerapose}, MV\mbox{-}RoMa outperforms existing baselines across all datasets and thresholds. These results demonstrate that MV\mbox{-}RoMa provides reliable matches in challenging low-texture environments and achieves strong performance under sparse sampling conditions.

\begin{table*}[h]
    \caption{\textbf{Multi-view camera pose estimation.}
    AUC of pose error (\%) at 3$^\circ$, 5$^\circ$, and 10$^\circ$ on the Texture-Poor SfM and IMC PhotoTourism datasets. 
    \textbf{Bold} and \underline{underlined} scores indicate the best and second-best results, respectively.}

    \centering
    \resizebox{0.95\textwidth}{!}{
    \setlength\tabcolsep{6pt} %
    \begin{tabular}{cccccccc} 
    \toprule
    \multirow{2}{*}{Type} & \multirow{2}{*}{Method}     &  \multicolumn{3}{c}{Texture-Poor SfM Dataset}  & 
    \multicolumn{3}{c}{IMC Dataset}
    \\ 
    \cmidrule(lr){3-5}
    \cmidrule(lr){6-8}  &     & AUC@3$\degree$ & AUC@5$\degree$ & AUC@10$\degree$ & AUC@3$\degree$ & AUC@5$\degree$ & AUC@10$\degree$ \\ 
    \midrule
    \multirow{5}{*}{Detector-Based} &   COLMAP~(SIFT+NN)~\cite{colmap} &  2.87 & 3.85 & 4.95 & 23.58 & 23.66 & 44.79 \\
    
    &   SIFT~\cite{sift} + NN + PixSfM~\cite{pixsfm}  & 3.13 & 4.08 & 5.09 & 25.54 & 34.80 & 46.73 \\
    &   D2Net~\cite{d2net} + NN + PixSfM & 1.54 & 2.63 & 4.54 & 8.91 & 12.26 & 16.79 \\
    
    &   R2D2~\cite{r2d2} + NN + PixSfM  & 3.79 & 5.51 & 7.84 & 31.41 & 41.80 & 54.65 \\

    &   SP~\cite{superpoint} + SG~\cite{superglue} + PixSfM & 14.00 & 19.23 & 24.55 & 45.19 & 57.22 & 70.47 \\
    
    \hline
    \multirow{5}{*}{Deep} 
    & VGGSfM~\cite{vggsfm}  & 1.86 & 6.26 & 21.10 & 45.23 & 58.89 & 73.92  \\
    & VGGT~\cite{vggt}  & 1.66 & 5.25 & 14.71 & 39.23 & 52.74 & 71.26 \\
    & Pi3~\cite{pi3} & 2.76 & 10.05 & 32.51 & 43.34 & 57.57 & 73.35 \\
    & Map-Anything~\cite{mapanything}  & 0.32 & 1.89 & 10.62 &  18.71 &  31.63 & 50.52 \\
    & Depth-Anything v3~\cite{dav3}  & 2.83 & 9.65 & 31.33 & 44.15 & 58.54 & 74.21 \\
    \hline
    \multirow{1}{*}{Point-Based} 
    & Mast3r-SfM~\cite{mast3r-sfm} & 7.10 & 18.08 & 42.44 & 31.77 & 46.36 & 64.37 \\    

    \hline    
    \multirow{4}{*}{Dense Matching} 
    & LoFTR~\cite{loftr} + PixSfM  & 20.66 & 30.49 & 42.01 & 44.06 & 56.16 & 69.61  \\
    & LoFTR + DFSfM~\cite{detectorfreesfm} & 26.07 & 35.77 & 45.43 & 46.55 & 58.74 & 72.19  \\
    & RoMa + Dense-SfM & \underline{49.94} & \underline{66.23} & \underline{81.41} & \underline{48.48} & \underline{60.79} & \underline{73.90} \\
    & MV-RoMa & \textbf{51.79} & \textbf{66.77} & \textbf{81.74} & \textbf{51.31} & \textbf{62.92} & \textbf{75.92} \\

    \bottomrule
    \end{tabular}
    }
    \label{tab:exp_camerapose}
    \vspace{-0.3cm}
\end{table*}
\subsection{Ablation Study}
\label{sec:ablation}
 
We ablate our key design choices on the ETH3D dataset using accuracy and completeness (density) of reconstructed 3D points, and on the HPatches dataset using DLT-estimated homography. Specifically, we report accuracy and completeness at varying distance thresholds for ETH3D, and AUC of the four image corners for HPatches.

\paragraph{Architecture Analysis.}
We assess the contribution of the multi-view encoder and multi-view matching refiner on ETH3D by comparing model variants with and without each component (Tab.~\ref{tab:ablation_arch}). 
The configuration without either component corresponds to the original RoMa~\cite{roma}.

The multi-view encoder (MV-Encoder in Tab.~\ref{tab:ablation_arch}) equips DINOv2 backbone with a track-guided module that produces multi-view-aligned features, while the multi-view matching refiner (MV-Refiner in Tab.~\ref{tab:ablation_arch}) replaces RoMa's original refinement module with our multi-view refiner.
Adding the MV-Encoder alone already improves over the RoMa baseline, and further incorporating the MV-Refiner yields the best performance, confirming that the two components contribute complementarily.

\begin{table}[t]
\caption{\textbf{Ablation on MV-Encoder and MV-Refiner.}
3D triangulation results on ETH3D for model variants with different component combinations.}
\small
\centering
\setlength{\tabcolsep}{4pt}
\begin{tabular}{cc|
>{\centering\arraybackslash}m{0.8cm}
>{\centering\arraybackslash}m{0.8cm}
>{\centering\arraybackslash}m{0.8cm}
>{\centering\arraybackslash}m{0.8cm}
}
\toprule

\multicolumn{2}{c|}{Components} & \multicolumn{2}{c}{Accuracy} & \multicolumn{2}{c}{Completeness} \\
\cmidrule(lr){1-2}\cmidrule{3-4}\cmidrule{5-6}
MV-Encoder & MV-Refiner &  1cm & 2cm &  1cm & 2cm \\
\midrule
           &            & 75.58 & 86.25 & 5.64 & 15.73 \\
\checkmark &            & 77.80 & 87.38 & 6.75 & 17.04 \\
\checkmark & \checkmark & \textbf{81.91} & \textbf{90.45} & \textbf{7.52} & \textbf{18.50} \\ 
\bottomrule
\end{tabular}

\label{tab:ablation_arch}
\vspace{-3mm}
\end{table}

\paragraph{Impact of Priors and Multi-View Interaction.}

We analyze the impact of geometric priors and multi-view interactions in Tab.~\ref{tab:ablation2}. 
First, we study the effect of the number of track tokens (Tab.~\ref{tab:ablation2}.A). Using a sufficient number of track tokens yields noticeably better homography accuracy than sparser configurations, and the performance degrades significantly when only a single track token is used. 
This shows that employing a sufficient set of track tokens enables effective multi-view feature exchange and leads to more accurate matching.
Second, we assess the benefit of joint multi-view processing by comparing our multi-view matching pipeline (1-to-5) with a pairwise (1-to-1) variant (Tab.~\ref{tab:ablation2}.B), where the 1-to-5 interaction is replaced by five independent 1-to-1 pairs. The lower accuracy in the pairwise setup shows that processing multiple views jointly yields better performance than handling each pair independently.

\begin{table}
    \caption{\textbf{Impact of multi-view priors and interactions.} \textbf{(A)} The number of track tokens. \textbf{(B)} The number of interacting views.}
    \small
    \centering    
    \begin{tabular}{l ccc}
    \toprule
    \multirow{2}{*}{Setup} & \multicolumn{3}{c}{AUC-DLT} \\ 
    \cmidrule(lr){2-4} 
          & @ 1px  & @ 3px  & @ 5px  \\          
     \midrule
    \multicolumn{4}{l}{\textit{A. Number of Track Tokens (V=5)}} \\ 
     \midrule

    \quad 512 (Full)  & \textbf{41.8} & \textbf{{69.9}} & \textbf{78.8} \\
    \quad 256 & \textbf{41.8} & 69.8 & 78.7 \\
    \quad 64 & \textbf{41.8} & \textbf{69.9} & 78.6 \\
    \quad 16 & 41.6 & 69.7 & 78.7 \\
    \quad 1 & 36.1 & 66.6 & 76.6 \\
    \midrule
    \multicolumn{4}{l}{\textit{B. Number of Interacting Views (T=512)}} \\ 
     \midrule

    \quad 5 (Full) & \textbf{41.8} & \textbf{{69.9}} & \textbf{78.8} \\ 
    \quad 1 (Pairwise) & 41.2 & 69.7 & 78.7 \\
     \bottomrule
    \end{tabular}

    \label{tab:ablation2}
    \vspace{-3mm}
\end{table}

\section{Conclusion}
We present MV\mbox{-}RoMa, a multi-view dense matching model that jointly estimates consistent dense correspondences across multiple images. Our efficient multi-view feature exchange architecture leverages geometric priors through track tokens and multi-view pixel-wise attention, enabling more accurate correspondence estimation than existing pairwise matchers. Extensive experiments show that MV\mbox{-}RoMa outperforms state-of-the-art methods across diverse downstream tasks and yields substantially denser and more accurate 3D reconstructions in the SfM pipeline.

\section{Acknowledgement}{
This work was supported by Samsung Advanced Institute of Technology (SAIT), Samsung Electronics (System LSI and Memory Division), IITP/NRF grants funded by the Korean government (MSIT, 2021-0-00105 Development of Model Compression Framework for Scalable On-Device AI Computing on Edge Applications, 2021-0-01343 Artificial Intelligence Graduate School Program, Seoul National University), and Inter-university Semiconductor Research Center (ISRC), SNU.
}

{
    \small
    \bibliographystyle{ieeenat_fullname}
    \bibliography{main}
}

\clearpage
\setcounter{page}{1}
\maketitlesupplementary
\setcounter{section}{0} 
\renewcommand\thesection{\Alph{section}}

\section{Grouping Images for SfM}
In this section, we detail how we partition the full set of images into groups for the SfM pipeline, as in the multi-view reconstruction experiments in Sec.~4.3 and Sec.~4.4 of the main paper.
Generally, SfM pipelines are formulated over large, unordered collections of images, whereas \textbf{MV-RoMa} processes only a small group at a time: one source image and up to $K$ target images. Given such a group, the model predicts dense correspondences from the source to all target images, which are then aggregated into multi-view tracks and fed into the downstream SfM pipeline for 3D reconstruction (Sec.~3.5). Since it is computationally infeasible to run our model on all possible image groups in a scene, we need to select a set of groups that provide informative tracks for SfM.

Ideally, each group should include images that share sufficient scene overlap so that the model can generate abundant reliable correspondences. In addition, dense correspondences should be available in both directions ($I_a \rightarrow I_b$ and $I_b \rightarrow I_a$) for any image pair, to enable bidirectional consistency check (Sec.~3.5). To satisfy these requirements, we adopt a \textit{two-stage sampling} scheme.
In the first stage, we construct an initial set of image groups, under a fixed group budget $G_{\text{budget}}$ (\ie the number of groups we are allowed to construct), where images within each group have strong visual overlap. 
Then, in the second stage, we add a minimal number of extra groups so that every image pair has correspondences in both directions.

\subsection{Sampling Procedure}

First, we compute a pairwise overlap matrix $O \in [0,1]^{M \times M}$ over all images, where each entry $o_{ij}$ is an \textit{overlap score} between images $i$ and $j$. We consider two simple strategies to define this score.

\paragraph{Visibility-based overlap.}
In our default setting, $O$ is obtained from an off-the-shelf pairwise matcher. Let $\mathcal{P}_i$ denote the set of all 2D pixel locations in image $i$, and let $\mathcal{N}_{ij} \subseteq \mathcal{P}_i$ be the subset of pixels in image $i$ that have a valid match in image $j$. Valid matches are determined from the matcher output, for example by thresholding a confidence score with a parameter $\tau_{\text{conf}}$. We then define
\begin{equation}
    o_{ij}
    \;=\;
    \frac{\lvert \mathcal{N}_{ij} \rvert}{\lvert \mathcal{P}_i \rvert}
\end{equation}
so that $o_{ij}$ measures the fraction of pixels in image $i$ that have a valid correspondence in image $j$.

\begin{figure}[t]
    \centering
    \includegraphics[
        width=\linewidth,
        trim={0.7cm 0 0.1cm 0},
        clip
    ]{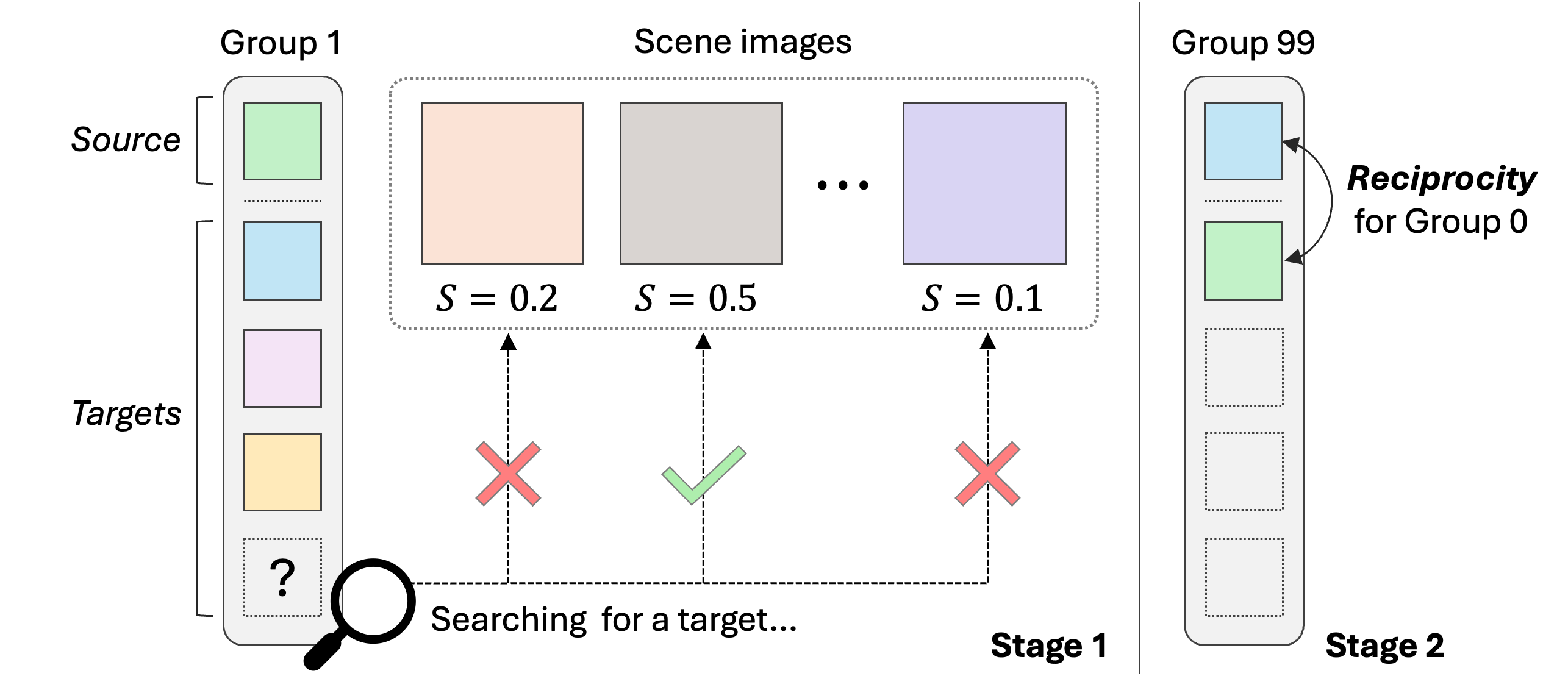}
    \caption{\textbf{Group Sampling Procedure.} We use a two-stage strategy: \textbf{Stage 1} greedily selects targets based on selection scores, while \textbf{Stage 2} generates additional groups to enforce reciprocity.}
    \label{fig:intro}
\vspace{-3mm}
\end{figure}

\paragraph{Feature-based overlap.}
In the setting where we only rely on image-level features, we instead define $O$ based on global descriptors from image retrieval models~\cite{netvlad, megaloc, cnn_retrieval}. Specifically, let $f_i$ and $f_j$ denote global descriptors for images $i$ and $j$, respectively. We then define
\begin{equation}
    o_{ij}
    \;=\;
    \frac{\langle f_i, f_j \rangle}{\lVert f_i \rVert \,\lVert f_j \rVert}
\end{equation}
\ie $o_{ij}$ is the cosine similarity between the two descriptors, providing an appearance-based measure of overlap with low computational cost.

\paragraph{Source quotas for balanced sampling.}
Given the overlap matrix $O$, we first determine how often each image is allowed to serve as a source. Under a limited group budget $G_{\text{budget}}$, we prioritize using images that exhibit strong overlap with many others as sources, since they are likely to produce more useful matches. 
At the same time, we aim to prevent certain images from being repeatedly overused while ensuring that all images serve as sources at least once.
To this end, we assign each candidate source image $i$ a quota $g_i$ that determines how many groups it can serve as a source for.
Let $N_i$ denote the number of high-overlap neighbors of image $i$ above a fixed threshold $\tau$, defined as:
\begin{equation}
    N_i
    \;=\;
    \sum_{j=1}^{M}
    \mathbb{I}\!\left[o_{ij} > \tau\right]
\end{equation}
We then set
\begin{equation}
    g_i \propto (N_i + 1)^{\beta}
\end{equation}
with an exponent $0 < \beta < 1$ (we use $\beta = 0.75$ by default). The additive term $+1$ guarantees that every image can act as a source at least once. In practice, we sample source images in proportion to $g_i$ and keep forming groups until we reach the global budget $G_{\text{budget}}$.

\paragraph{Greedy group construction.}
Once a source image $i$ is selected to construct a group, we add target images incrementally. At each step, we use a \textit{selection score} that (i) favors strong overlap with the source, (ii) encourages coherence with already selected targets, and (iii) discourages repeated use of the same ordered pair $(i,j)$. Let $\mathcal{T}_{\text{cur}}$ denote the current set of target images. For a candidate target image $j \notin \mathcal{T}_{\text{cur}}$, we define the selection score as:
\begin{equation}
    \mathrm{Score}(i, j \mid \mathcal{T}_{\text{cur}})
    \;=\;
    \Big(
        \alpha_{\text{src}}\, o_{ij}
        +
        \alpha_{\text{tgt}}
        \sum_{k \in \mathcal{T}_{\text{cur}}} o_{kj}
    \Big)
    \cdot
    \frac{1}{1 + \lambda\, c_{ij}}
\end{equation}
where $\alpha_{\text{src}}, \alpha_{\text{tgt}}, \lambda \ge 0$ are scalar hyperparameters, and $c_{ij} \in \mathbb{N}$ counts how many times the directed pair $(i,j)$ has already been selected in previously constructed groups. The linear term in parentheses measures how well $j$ fits the current group: $o_{ij}$ captures alignment with the source, and $\sum_{k \in \mathcal{T}_{\text{cur}}} o_{kj}$ captures alignment with the existing targets, balanced by $\alpha_{\text{src}}$ and $\alpha_{\text{tgt}}$. The multiplicative factor $(1 + \lambda c_{ij})^{-1}$ applies a soft penalty to repeated use of the same pair, encouraging the sampler to favor a more diverse set of image pairs. At each step, we evaluate this score for all remaining candidates $j \notin \mathcal{T}_{\text{cur}}$ and select the image with the highest score, until we reach the maximum group size (up to $K$ targets) or no candidates remain.

\paragraph{Group augmentation for reciprocity.}
To support the bidirectional consistency check in Sec.~3.5, we add a second stage that ensures all image pairs are bidirectional by creating additional groups.
Whenever a directed pair $(i,j)$ is created during the first stage (\ie, $i \rightarrow j$), we record that $j$ needs to connect back to $i$. For each image $j$ that still has such remaining reciprocity requirements, we create additional groups with $j$ as a source image. In these groups, we first fill the target slots with images $k$ that still require reciprocity with $j$. 
For remaining target slots, we add further targets chosen using the same selection score as above, but restricting candidates to images that have already been paired with $j$ (either $j \rightarrow k$ or $k \rightarrow j$). This procedure avoids creating new one-sided pairs in the second stage while ensuring that all image pairs have dense correspondences in both directions.

\section{Runtime and Efficiency}

\subsection{MV-RoMa's Runtime}
We compare the inference time of MV-RoMa against RoMa at 448×448 resolution on an NVIDIA RTX 6000 Ada GPU. 
RoMa requires 85 ms for a single source–target pair (1-to-1) with a batch size of 1, which scales to 340 ms when processing five pairs in parallel with a batch size of 5. In contrast, MV-RoMa processes one source image jointly with five target images (1-to-5) in 280 ms.

\subsection{SfM Runtime}
In our experiments in the main paper, we set the group budget $G_{\text{budget}}$ to around $N\sqrt{N}$ , ensuring $\mathcal{O}(N\sqrt{N})$ complexity. ($N$ is the number of images in a scene). We use $\sqrt{N}$ to reflect scene scale; all our evaluation scenes contain fewer than $100$ images, and for large $N$, a constant value can replace $\sqrt{N}$. Within each group, matching for track clustering (via UFM) is performed $K$ times (source-to-targets, where $K$ is the number of target images) rather than combinatorially. 

\begin{table}[h]
    \centering
    \caption{Runtime comparison of each method for SfM reconstruction on a 20-image scene from the Texture-Poor SfM dataset.}
    \label{tab:tab_sfm_runtime}
    \resizebox{\columnwidth}{!}{
        \begin{tabular}{c|c|ccccc}
        \toprule
        Method & Budget & Match & Select \& Filter & COLMAP & Refine & Total \\
        \midrule
        DF-SfM w/ RoMa &  - & 1.5min & 2sec & 2.5min  & 1.5min & 5.5min \\
        \midrule
       \multirow{2}{*}{MV-RoMa (Ours)}& Full & 3min & 3sec & 3min  & - &  6min  \\
         & Half & 1.5min & 2sec & 2min & - &  3.5min \\
        \bottomrule
        \end{tabular}
    }
\vspace{-3mm}
\end{table}

To demonstrate the efficiency of MV-RoMa, Tab.~\ref{tab:tab_sfm_runtime} reports runtimes on the sub-scene \textit{'20bag000'}(containing 20 images) from scene \textit{1000} of the Texture-Poor SfM dataset, comparing MV-RoMa with DF-SfM+RoMa~\cite{detectorfreesfm} using a single A6000 GPU.
For DF-SfM+RoMa, we use exhaustive pairs following DF-SfM's evaluation protocol and additionally apply bidirectional consistency filtering (Sec.3.5.1) for fair comparison. Our \textit{Half Budget} configuration (from 60 to 70 groups with each group containing 1-to-4 pairs), which reduces the group budget from $N\sqrt{N}$ to $\frac{1}{2}N\sqrt{N}$, achieves a total runtime of 3.5 minutes compared to DF-SfM's 5.5 minutes. This is mainly because MV-RoMa does not require track refinement performed in the DF-SfM pipeline, saving post-processing time after COLMAP reconstruction. 
As for per-group runtime, RoMa takes 0.25 seconds per 1-to-1 pair, while MV-RoMa takes 1.4 seconds per 1-to-4 group (including 0.25 seconds for running UFM per group). Note that the time cost for match selection and filtering (Sec.3.5.1) is negligible.

\begin{table}[ht!]
    \centering
    \caption{Performance comparison between DF-SfM and MV-RoMa on the scene \textit{1000} of the Texture-Poor SfM dataset.}
    \label{tab:tab_full_half_performance}
    
    \resizebox{0.99\columnwidth}{!}{
        \begin{tabular}{c|c|c}
            \toprule
            Method & Budget & Texture-Poor AUC@3$^\circ$ / 5$^\circ$ / 10$^\circ$  \\ 
            \midrule
            DF-SfM w/ RoMa & - & 45.22 / 63.54 / 80.54 \\ 
            \midrule
            \multirow{2}{*}{MV-RoMa (Ours)} & Full &\textbf{60.71} / \textbf{74.66} / 85.88 \\
                                            & Half &  59.97 / 74.60 / \textbf{86.23} \\     
            \bottomrule
        \end{tabular}
    }
    \label{tab:compare_dfsfm_texturepoor}
\end{table}

To evaluate the relationship between group budget for matching and pose accuracy, we compare camera pose estimation with full and half budgets in scene $1000$ from the Texture-Poor SfM. As shown in Tab.~\ref{tab:compare_dfsfm_texturepoor}, halving the budget results in only a marginal drop in accuracy, demonstrating that MV-RoMa's gains mainly stem from superior multi-view track consistency rather than high density of image pairs.

\section{Training Data Construction}

\paragraph{MegaDepth.}
For MegaDepth~\cite{megadepth}, we use the multi-view image groups provided by DF-SfM~\cite{detectorfreesfm}, which already ensure sufficient co-visibility within each group. From each group, we randomly select one source image and four target images.

\paragraph{ScanNet.}
For ScanNet~\cite{scannet}, we build new multi-view groups from the two-view training split of LoFTR~\cite{loftr}. First, we derive candidate source--target pairs from LoFTR’s training pairs and compute dense correspondences using ground-truth depth maps, camera poses, and intrinsics via geometric warping and verification. 
For each pair, we compute the \emph{overlap ratio} as the ratio of the number of valid correspondences to the total number of pixels sampled for correspondence estimation in the source image, and keep only pairs with overlap ratio in the range $[15\%, 80\%]$ to discard both nearly disjoint and almost identical views.
Next, to enforce view diversity among target images from the same source image, we compute pairwise overlap between every pair of candidate target images, similar to the overlap ratio applied between source and target.
Whenever two targets overlap by more than $70\%$, we randomly remove one of them. From the remaining candidates, we form training groups by selecting four targets per source, ensuring that each frame appears at most once in the training set. This procedure yields multi-view groups with sufficient geometric overlap for learning, while avoiding redundant or overly similar frame combinations.

\section{Analysis of Track Token}
\subsection{Flexibility Across Matchers}

\begin{table}[h]
\centering
\caption{Homography estimation on the HPatches dataset using different off-the-shelf matchers to generate track tokens.}
\label{tab:tab_homography_matcher}
\resizebox{\columnwidth}{!}{
  \begin{tabular}{lcc}
    \toprule
    \multirow{2}{*}{Matcher model} & \multicolumn{2}{c}{AUC @1px / @3px / @5px} \\
    \cmidrule(lr){2-3}
     & DLT & RANSAC \\
    \midrule
    ALIKED~\cite{aliked} + LightGlue~\cite{lightglue} 
      & 42.9 / 70.3 / 79.1 & 42.9 / 71.0 / 80.2 \\
    UFM~\cite{ufm} 
      & 41.8 / 69.9 / 78.8 & 42.6 / 71.5 / 81.0 \\
    \bottomrule
  \end{tabular}
}
\vspace{-3mm}
\end{table}

As discussed in Sec.~3.3 of the main paper, our framework can use any off-the-shelf matcher model to generate track tokens. To demonstrate this flexibility, we train and evaluate our model using ALIKED~\cite{aliked} and LightGlue~\cite{lightglue} as the detector and matcher, respectively, applying the same clustering-based sampling strategy on the resulting matches to construct track tokens.
Tab.~\ref{tab:tab_homography_matcher} shows that using ALIKED+LightGlue achieves homography estimation performance comparable to using UFM~\cite{ufm}. This indicates that keypoint-based models can also provide effective geometric priors through track token construction.

\subsection{Effect of Clustering-based Sampling}

\begin{table}[h]
\centering
\caption{Homography estimation on the HPatches dataset, comparing MV-RoMa's performance under clustering-based sampling versus random sampling.}
\label{tab:tab_homography_lightglue}

\resizebox{\columnwidth}{!}{
  \begin{tabular}{lcccc}
    \toprule
    \multirow{2}{*}{Matcher model} & \multirow{2}{*}{Clustering} & \multirow{2}{*}{Error (median)} & \multicolumn{2}{c}{AUC @1px / @5px} \\
    \cmidrule(lr){4-5}
     &  &  & DLT & RANSAC \\
    \midrule
    \multirow{2}{*}{ALIKED~\cite{aliked} + LG~\cite{lightglue}} 
      &         & \textbf{0.50} & 42.8 / 78.9 & \textbf{43.1} / 80.1 \\
      & \checkmark & \textbf{0.50} & \textbf{42.9} / \textbf{79.1} & 42.9 / \textbf{80.2} \\
    \midrule
    \multirow{2}{*}{UFM~\cite{ufm}} 
      &        & 0.53 & 41.5 / \textbf{78.8} & 42.4 / 80.7 \\
      & \checkmark & \textbf{0.51} & \textbf{41.8} / \textbf{78.8} & \textbf{42.6} / \textbf{81.0} \\
    \bottomrule
  \end{tabular}
}
\vspace{-3mm}
\end{table}

\begin{figure}[h]
    \centering
    \includegraphics[
        width=\linewidth,
    ]{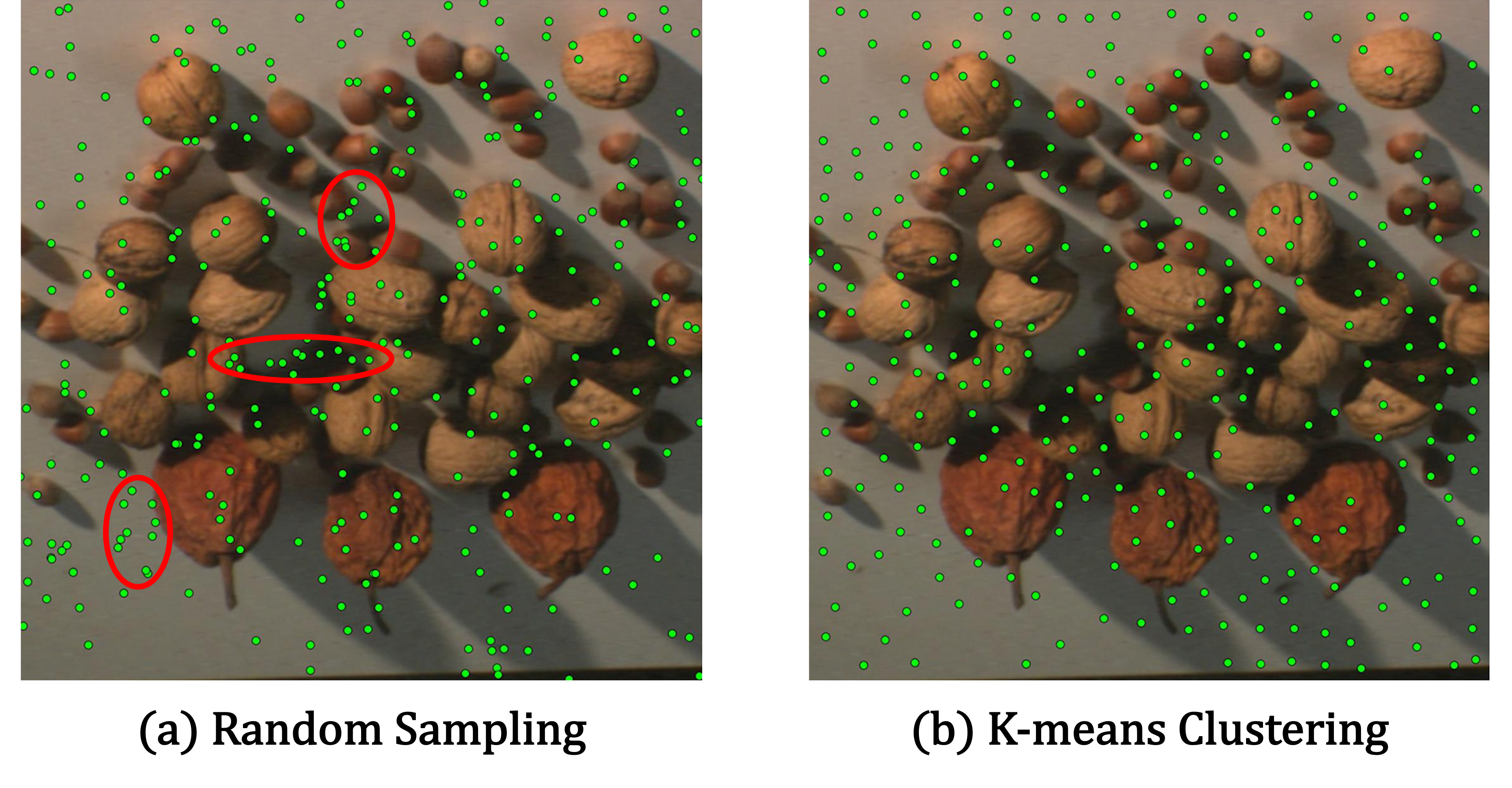}
    \vspace{-5mm}
    \caption{\textbf{Random vs Clustering.} \textbf{(a)} Random sampling results in spatially uneven distributions with redundant aggregations (red circles). \textbf{(b)} Our clustering-based sampling approach generates a spatially uniform distribution of track tokens, ensuring better coverage for feature exchange.}
    
    \label{fig:random_vs_clustering}
\vspace{-3mm}
\end{figure}

As explained in Sec.~3.3 of the main paper, we construct track tokens by applying clustering-based sampling to the initial matching results. As illustrated in Fig.~\ref{fig:random_vs_clustering}, our k-means clustering yields a more spatially uniform distribution of tracks, whereas random sampling often produces highly aggregated keypoints in certain regions (red circles in Fig.~\ref{fig:random_vs_clustering}.a). Such uneven coverage can leave substantial regions without any tracks, reducing the chance of retrieving information there and leading to suboptimal results.
As reported in Tab.~\ref{tab:tab_homography_lightglue}, clustering-based sampling outperforms random sampling in homography estimation.

\subsection{Effect of Spatial Coverage}
\begin{table}[h]
\centering
\caption{Effect of reduced spatial coverage on homography estimation on HPatches.}
\label{tab:tab_homography_coverage}
\resizebox{\columnwidth}{!}{
  \begin{tabular}{lcc}
    \toprule
    \multirow{2}{*}{Border Masked} & \multicolumn{2}{c}{AUC @1px / @3px / @5px} \\
    \cmidrule(lr){2-3}
     & DLT & RANSAC \\
    \midrule
    5\%  & 41.9 / 69.9 / 78.8 & 42.5 / 71.3 / 80.7 \\
    10\% & 41.8 / 69.9 / 78.8 &  42.2 / 71.3 / 80.8 \\
    15\% & 41.4 / 69.9 / 78.9 & 41.7 / 70.7 / 80.2 \\
    \bottomrule
  \end{tabular}
}
\vspace{-3mm}
\end{table}

\begin{figure}[h]
    \centering
    \includegraphics[
        width=\linewidth,
        trim={0 1.0cm 0 0cm},
    ]{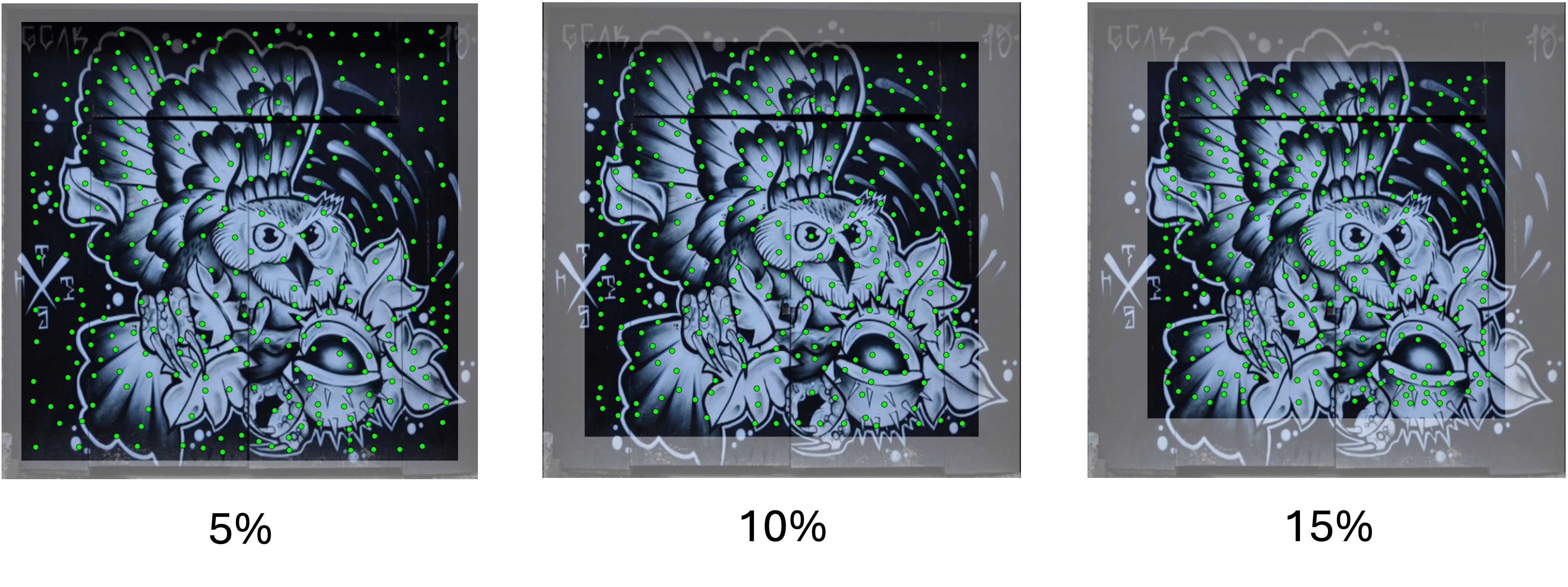}
    \caption{\textbf{Spatial Coverage Analysis.} We mask the outer 5\%, 10\%, and 15\% border regions (bright area) while maintaining a fixed total number of track tokens ($T=512$, only 300 tokens are shown for visualization). The experiment confirms that well-distributed tokens are essential for optimal results.}
    \label{fig:spatial_coverage}
\vspace{-3mm}
\end{figure}

To further validate the importance of spatial coverage, we conduct controlled experiments that artificially restrict where track tokens can be generated. As illustrated in Fig.~\ref{fig:spatial_coverage}, we progressively suppress track tokens in the outer 5\%, 10\%, and 15\% border regions, while keeping the total number of track tokens fixed at 512. As reported in Tab.~\ref{tab:tab_homography_coverage}, performance degrades as spatial coverage is reduced, confirming that well-distributed track tokens across the entire image are essential for optimal results. This finding highlights the importance of clustering-based sampling, which naturally promotes 
a more uniform spatial distribution of track tokens


\end{document}